\documentclass[letterpaper, 10pt, conference]{ieeeconf}  

\IEEEoverridecommandlockouts                              

\usepackage{amsmath}
\usepackage{amssymb}
\usepackage{amsfonts}

\usepackage{amsthm}
\usepackage{bm}
\usepackage{graphicx}
\usepackage[dvipsnames]{xcolor}
\usepackage{threeparttable}
\usepackage{diagbox}
\usepackage{cite}
\usepackage{array}
\usepackage{hyperref}

\usepackage{balance}
\usepackage{comment}

\newcommand\norm[1]{\left\lVert#1\right\rVert}

\newcolumntype{C}[1]{>{\centering\arraybackslash}p{#1}}

\newcommand{\figref}[1]{\mbox{Fig.~\ref{#1}}}
\newcommand{\tabref}[1]{\mbox{Table~\ref{#1}}}

\newcommand{\secref}[1]{\mbox{Section~\ref{#1}}}

\newtheorem*{proposition}{Proposition}
\theoremstyle{remark}
\newtheorem*{remark}{Remark}

\title{\LARGE \bf
Adaptive Step Duration for Accurate Foot Placement: Achieving Robust Bipedal Locomotion on Terrains with Restricted Footholds
}
\author{Zhaoyang Xiang$^{1}$, Victor Paredes$^{1}$, Guillermo A. Castillo$^{2}$, and Ayonga Hereid$^{1}$
\thanks{*This work was supported in part by the National Science Foundation under grant FRR-21441568. }%
\thanks{$^{1}$Mechanical and Aerospace Engineering, The Ohio State University, Columbus, OH, USA. {\tt\footnotesize (xiang.295, paredescauna.1, hereid.1)@osu.edu.}}%
\thanks{$^{2}$Electrical and Computer Engineering, The Ohio State University, Columbus, OH, USA;  {\tt\footnotesize castillomartinez.2@osu.edu}}
}

\begin{document}

\maketitle

\begingroup
\renewcommand\thefootnote{}
\footnotetext{This work was accepted and presented at the IEEE/RSJ International Conference on Intelligent Robots and Systems (IROS 2025). DOI: 10.1109/IROS60139.2025.11246171.}
\addtocounter{footnote}{-1}
\endgroup

\begin{abstract}
Traditional one-step preview planning algorithms for bipedal locomotion struggle to generate viable gaits when walking across terrains with restricted footholds, such as stepping stones. 
To overcome such limitations, this paper introduces a novel multi-step preview foot placement planning algorithm based on the step-to-step discrete evolution of the Divergent Component of Motion (DCM) of walking robots. Our proposed approach adaptively changes the step duration and the swing foot trajectory for optimal foot placement under constraints, 
thereby enhancing the long-term stability of the robot and significantly improving its ability to navigate environments with tight constraints on viable footholds. We demonstrate its effectiveness through various simulation scenarios with complex stepping-stone configurations and external perturbations. These tests underscore its improved performance for navigating foothold-restricted terrains, even with external disturbances.
\end{abstract}

\section{Introduction}

\begin{figure}[t]
    \centering
    \vspace{2mm}
    \includegraphics[width=1\linewidth]{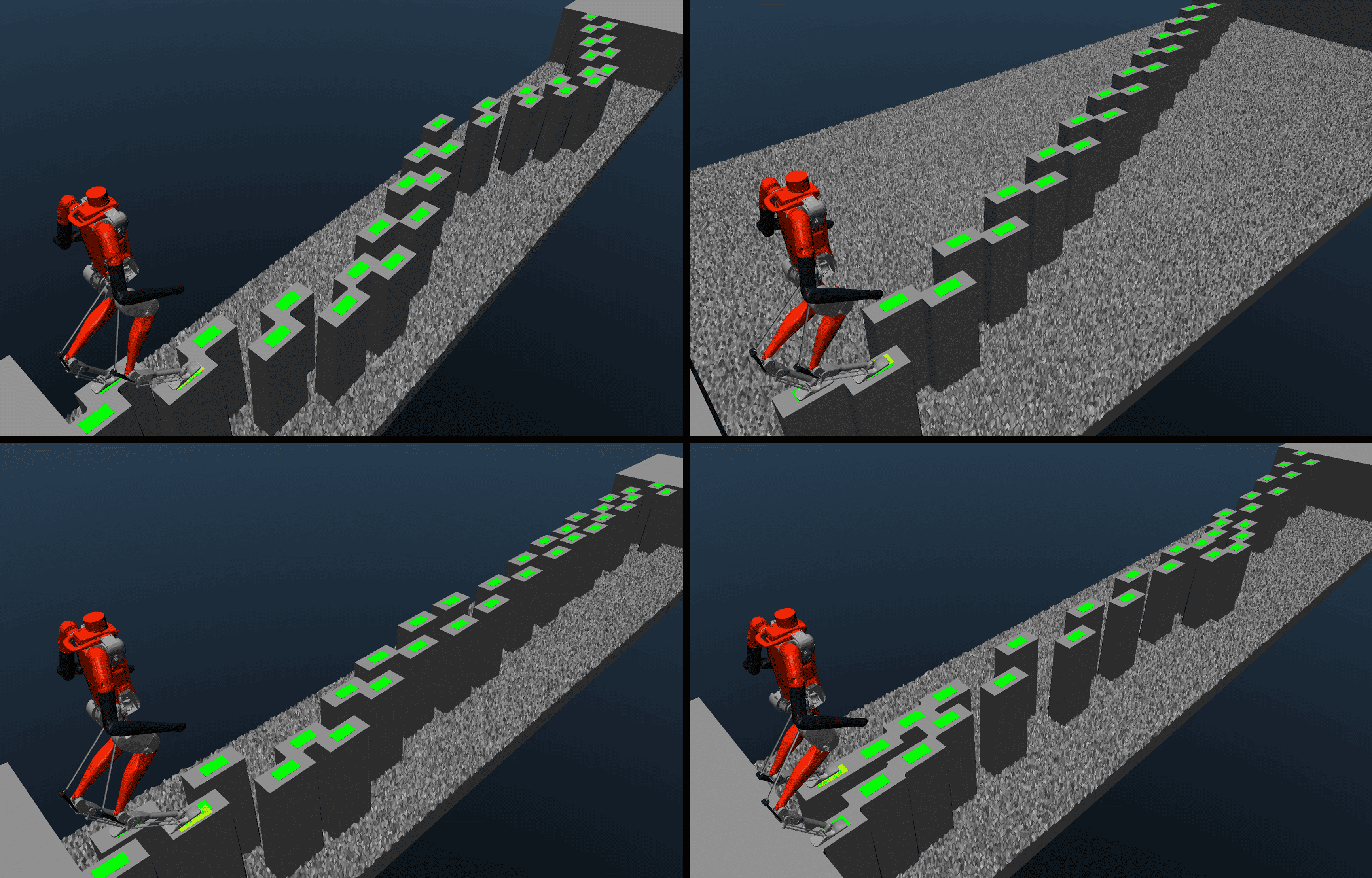}
    \caption{Digit robot stably walks across four challenging stepping stone scenarios in MuJoCo simulation by dynamically adjusting both step duration and stepping locations.}
    \label{fig:sim_overview}
    \vspace{-2mm}
\end{figure}

Bipedal robots are designed to collaborate with humans while navigating complicated and crowded environments built for human use. These environments often restrict the robot's viable footholds, requiring the robot to adaptively plan its gaits in real-time to ensure each footstep lands within feasible regions. In the literature on bipedal locomotion, such limited regions are often regarded as stepping stones, as illustrated in \figref{fig:sim_overview}. Many dynamic walking algorithms fail to successfully tackle stepping stones as they often assume the foot can be placed anywhere within the allowable workspace of the swing foot. 
A noticeable exception in the literature is the use of full-body trajectory optimization with foothold constraints. 
For example, a library of gaits is often generated offline with a set of fixed stepping-stone profiles, and then interpolated online given the real-time feedback of the next steppable area~\cite{nguyen2017dynamic} or adapted through an online MPC with a whole-body controller~\cite{li2023dynamic}. However, the adaptability of such approaches is limited due to the difficulty in optimizing a full-body trajectory in real-time for any given stepping stone profile. 

A popular approach to mitigate the complexity of robot dynamics in gait planning is to use reduced-order template models to represent the critical progression of bipedal robots during locomotion. The most commonly used model, the linear inverted pendulum model (LIPM), describes the center of mass (CoM) dynamics with an inverted pendulum with a constant height~\cite{kajita20013d,xiong2019orbit,tedrake2015closedform,sugihara20213d}. 
To provide a criterion for gait stability of pendulum-based CoM dynamics models, the capturability~\cite{koolen2012capturabilitybased} or the Divergent Component of Motion (DCM)~\cite{englsberger2015threedimensional} decouples the stable and unstable dynamics based on the viable states (i.e., states that do not lead to falling). These analyses also show that a variable step duration can increase the range of viable states and allow robots to prevent falling by taking faster and wider steps. However, such a time-varying variable usually leads to nonlinear and non-convex planning problems, which require heavy computing resources and thus impede its implementation in online controllers. Griffin \emph{et al.} formulate a continuous-time MPC for the time-varying DCM as a mixed-integer quadratically constrained quadratic program (MIQCP) that shows good resistance to force perturbations but suffers from slow online solving (average about 72 ms)~\cite{griffin2016model}.

The decoupling of the stable and unstable dynamics through DCM allows the gait planner to regulate both stepping position and timing conveniently. 
Khadiv \emph{et al.} introduce the DCM offset to represent the viability bounds of the LIPM model and formulate the desired step position and step duration into an online quadratic programming (QP)~\cite{khadiv2020walking}. Their foot placement planner finds an optimal foot placement to ensure bounded viable states, thus improving perturbation resistance. There are also extended works for more complicated stepping strategies~\cite{zhang2022stride} and more challenging walking perturbations~\cite{kim2023foot}, which implies their potential application in bipedal locomotion on stepping stones. Nonetheless, these approaches rely on one-step preview planning, in which the gait features are determined without considering the restrictions on future foothold locations. Consequently, this limits their adaptability to traverse randomly placed stepping stones.

In this work, we develop a multi-step preview gait planning approach to improve the overall stability and long-horizon viability of bipedal walking, even in the presence of severe constraints on the footholds. We introduce a multiple-step discrete Model Predictive Control (MPC) formulation to adjust step durations and stepping positions while walking over randomly located stepping stones. 
Our formulation uses the modified form of the DCM based on the contact angular momentum rather than the linear velocity of the base. This is inspired by the improved performance of the angular momentum-based linear inverted pendulum (ALIP) in bipedal locomotion on various terrains due to better approximation of robot dynamics~\cite{gong2022zero,gibson2022terrainadaptive,gao2023timevarying}. We validated the proposed approach with the Digit humanoid in simulation under various types of stepping stone profiles and perturbations. Compared with the one-step preview algorithm in ~\cite{khadiv2020walking}, our planner achieves more robust bipedal locomotion over challenging scenarios.


The remainder of the paper is organized as follows: \secref{sec:dcm_overview} introduces the modified DCM formulation, and \secref{sec:hlc} proposes a multi-step preview planner that adaptively changes step durations to remain stable on stepping stones. \secref{sec:llc} discusses the modification in the low-level task space tracking controller in response to adaptive step durations. \secref{sec:results} presents the simulation results using the Digit robot, and \secref{sec:conclusion} provides a brief conclusion, summarizing the contributions and discussing future plans.

\section{Divergent Component of Motion Analysis of Bipedal Locomotion}
\label{sec:dcm_overview}

This section introduces the modified DCM formulation based on the contact angular momentum-based linear inverted pendulum (ALIP) model~\cite{gong2022zero}. We also explicitly discuss the bounds of step positions and durations for stable locomotion.  

\subsection{Center of Mass Dynamics of Bipedal Robots}

We consider the ALIP model on flat ground with an underactuated single-support phase and an instantaneous double-support phase~\cite{gong2022zero}. Let $(x_c, y_c, z_c)$ be the CoM position in the right-handed contact foot frame, the ALIP dynamics are given by~\cite{gong2022zero}:
\begin{align} \label{eq:alip_dynamics}
\begin{split}
    \dot{x}_c &= \frac{\mathbf{L}^y}{mz_c},\quad\quad\ \dot{y}_c = -\frac{\mathbf{L}^x}{mz_c}, \\
    \dot{\mathbf{L}}^x &= -mgy_c,\quad \dot{\mathbf{L}}^y = mgx_c, 
\end{split}
\end{align}
where $z_c$ is the constant CoM height, $\mathbf{L}^x, \mathbf{L}^y$ are the $x, y$-component of the angular momentum about the contact point (i.e., contact angular momentum), $m$ is the mass of the robot, and $g$ is the gravitational acceleration. Compared to the LIPM described in~\cite{kajita20013d}, the ALIP model uses the contact angular momentum $\mathbf{L}^x, \mathbf{L}^y$ instead of linear velocities of the CoM. This modification reduces the mismatch between the template model and the real robot state, as it only assumes that the centroidal angular momentum oscillates about zero, rather than its rate of change being zero~\cite{gong2022zero}.

\begin{figure}[t]
    \centering
    \vspace{2mm}
    \includegraphics[width=\linewidth]{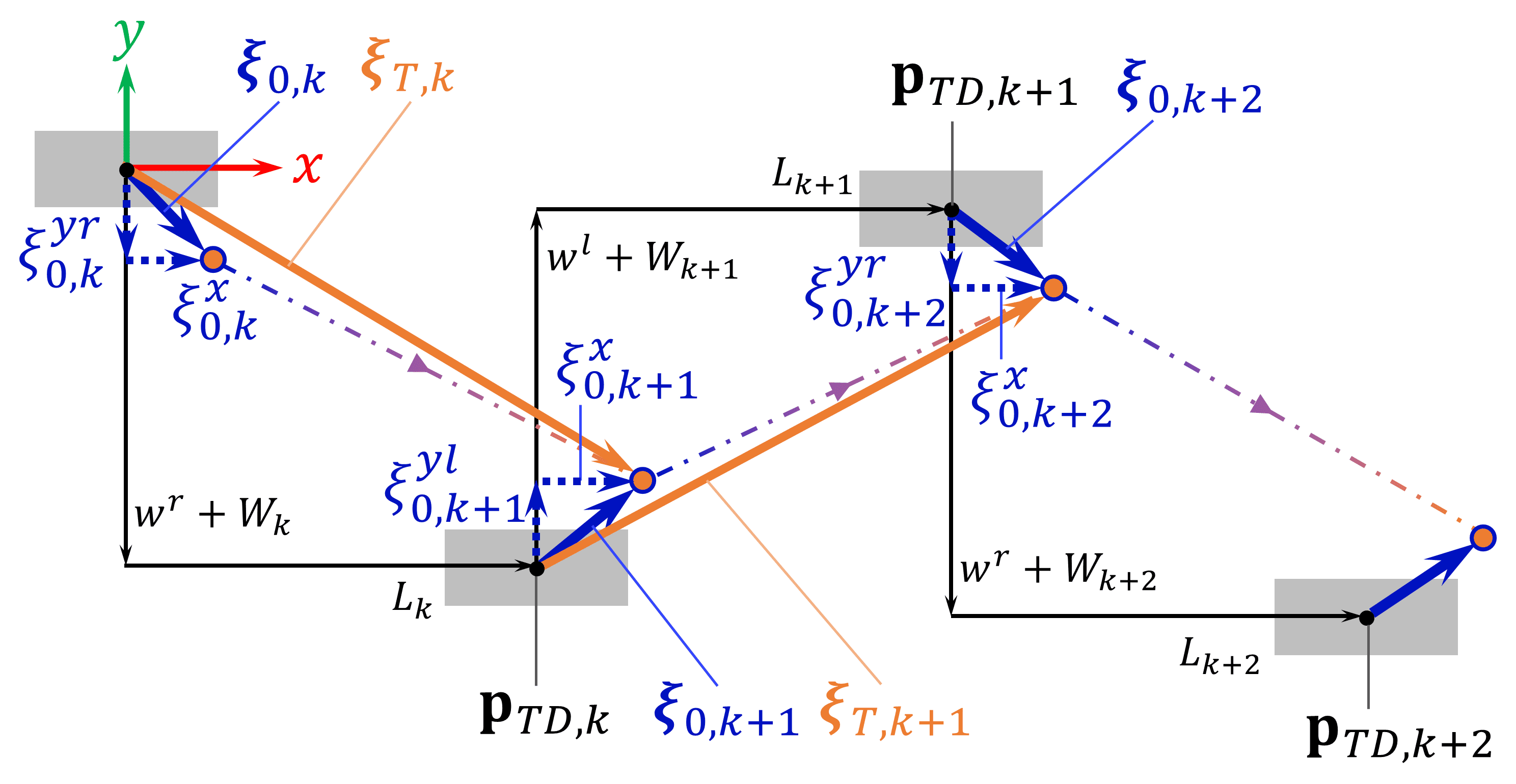}
    \caption{Schematic view of the DCM evolution with step positions and footprints. The blue arrows mark the initial DCM at the beginning of steps and the orange ones mark the final DCM. During the $k$-th step, the DCM evolves from $\boldsymbol{\xi}_{0,k}$ to $\boldsymbol{\xi}_{T,k}$ and is reset to $\boldsymbol{\xi}_{0,k+1}$ in the new contact frame after the touchdown.}
    \label{fig:footprints_and_dcm}
    \vspace{-2mm}
\end{figure}

\subsection{Discrete Dynamics of the initial DCM}

The ALIP dynamics in \eqref{eq:alip_dynamics} can be transformed into a form that decouples the stable and unstable components, with the latter referred to as the Divergent Component of Motion (DCM)~\cite{englsberger2015threedimensional}. Using the states of the ALIP model, the DCM $\boldsymbol{\xi}=[\xi^x, \xi^y]^T$ in the contact frame is given by
\begin{align} \label{eq:dcm_def}
    \boldsymbol{\xi} = \begin{bmatrix}
        x_c \\ y_c
    \end{bmatrix} + \frac{1}{\lambda} \frac{1}{mz_c} \begin{bmatrix}
        \mathbf{L}^y \\ -\mathbf{L}^x
    \end{bmatrix},\quad \lambda = \sqrt{\frac{g}{z_c}}.
\end{align}
where the normalized angular momentum terms replace velocity terms from LIPM.

Substituting the ALIP dynamics \eqref{eq:alip_dynamics} into \eqref{eq:dcm_def}, we can obtain the DCM dynamics as $\dot{\boldsymbol{\xi}} = \lambda \boldsymbol{\xi}$ with the solution
\begin{align} \label{eq:dcm_solution}
    \boldsymbol{\xi}(t) = \boldsymbol{\xi}_0 e^{\lambda t},\quad t\in[0,T],
\end{align}
where $t$ is the elapsed time (or relative time) from the beginning of the swing phase, $T$ is the step duration, and $\boldsymbol{\xi}_0$ is the initial DCM at $t=0$. During a walking step, the DCM $\boldsymbol{\xi}(t)$ increases exponentially over the time $t$, and is reset to another initial value after a swing foot touchdown due to the update of the contact frame (see \figref{fig:footprints_and_dcm}). For the $k$-th step ($k\in \mathbb{N^+}$), let $\boldsymbol{\xi}_{0,k}$ and $\boldsymbol{\xi}_{T,k}$ be the initial and final values of the DCM during the swing phase, and $\boldsymbol{\xi}_{0,k+1}$ be the initial value of the ($k$+1)-th step which is also referred to as the DCM offset at the end of the $k$-th step in~\cite{khadiv2020walking}. Then, the reset map of the initial DCM is given by
\begin{align} \label{eq:dcm_reset_map}
\begin{split}
    \boldsymbol{\xi}_{0,k+1} &= \mathit{\Delta}_k (\boldsymbol{\xi}_{0,k}) \\
    &= \boldsymbol{\xi}_{T,k} - \mathbf{p}_{TD,k} \\
    &= \boldsymbol{\xi}_{0,k} e^{\lambda T_k} - \mathbf{p}_{TD,k},
\end{split}
\end{align}
where $\mathbf{p}_{TD,k}=[p^x_{TD,k}, p^y_{TD,k}]^T$ is the $k$-th step position (i.e., Touch-Down position) in the current contact frame and $T_k$ is the $k$-th step duration.

Note that the DCM is innately decoupled in $x$- and $y$-direction. For a 3-D bipedal model, there will be an alternating step width $w^{l/r}$ for the left or right swing foot, even if the robot is stepping in place. Thus, we define the lateral step position $p^y_{TD,k}:= w^{l/r} + W_k$, where $w^{l/r}$ is a fixed step width of the left or right swing foot during in-place stepping, and $W_k$ is an offset to $w^{l/r}$ that accounts for the actual lateral movement. That is, given a fixed value $w>0$, the signed step width for the left swing foot is $w^l=w>0$ and for the right $w^r=-w<0$. For instance, \figref{fig:footprints_and_dcm} shows rightward walking in the $y$-direction with $W_k<0$.

For a periodic gait with a constant step position $\mathbf{p}_{TD,k}=[L, w^{l/r}+W]^T$ and duration $T_k=T$, the nominal value of the initial DCM $\boldsymbol{\xi}_{0,\mathrm{nom}}$ can be derived from \eqref{eq:dcm_reset_map} as
\begin{align} \label{eq:dcm_xy_periodic}
    \boldsymbol{\xi}_{0,\mathrm{nom}} = \begin{bmatrix}
        \xi^x_{0,\mathrm{nom}} \\
        \xi^{yl/r}_{0,\mathrm{nom}}
    \end{bmatrix} = \begin{bmatrix}
        \dfrac{L}{e^{\lambda T}-1} \\
        \dfrac{w^{l/r}}{e^{\lambda T}+1} + \dfrac{W}{e^{\lambda T}-1}
    \end{bmatrix},
\end{align}
where $\xi^{yl/r}_{0,\mathrm{nom}}$ of the left or right swing foot are solved by considering two consecutive steps (i.e., one cycle of lateral walking).

In the $x$-direction, given a bounded set $\mathcal{D}$ of allowable step position and $\mathcal{T}$ of allowable step duration, we can obtain a bounded set $\mathcal{X}$ of the nominal initial DCM computed by the values in $\mathcal{D}$ and $\mathcal{T}$ using \eqref{eq:dcm_xy_periodic}, i.e., $ \xi^x_{0,\mathrm{nom}}(L,T): \mathcal{D}\times \mathcal{T} \rightarrow \mathcal{X}$, $\forall L \in \mathcal{D}$, $T\in \mathcal{T}$. Each element in this set represents a particular periodic walking gait corresponding to a foot placement $(L,T)$, which leads to the following proposition.

\begin{proposition} \label{prop:dcm_boundedness}
    Given appropriate allowable sets $\mathcal{D}$, $\mathcal{T} \subset \mathbb{R}$ of the $k$-th step, for any initial DCM $\xi^x_{0,k}$ in the corresponding $\mathcal{X}$ that evolves once into $\xi^x_{0,k+1}$ using \eqref{eq:dcm_reset_map}, there exists at least one allowable foot placement $(L_k,T_k)$ such that $\xi^x_{0,k+1}$ still remains in $\mathcal{X}$.
\end{proposition}

\begin{proof}
    Let $\mathcal{D} = [L_{\mathrm{min}}, L_{\mathrm{max}}]$ with $L_{\mathrm{min}} < 0 < L_{\mathrm{max}}$ and $\mathcal{T} = [T_{\mathrm{min}}, T_{\mathrm{max}}]$ with $0< T_{\mathrm{min}} < T_{\mathrm{max}}$, the corresponding $\mathcal{X} = [\xi^x_{0,\mathrm{min}}, \xi^x_{0,\mathrm{max}}]$ is given by \eqref{eq:dcm_xy_periodic} as
    \begin{align}
    \begin{split}
        \xi^x_{0,\mathrm{min}} &= \frac{L_{\mathrm{min}}}{e^{\lambda T_{\mathrm{min}}}-1} < 0, \\
        \xi^x_{0,\mathrm{max}} &= \frac{L_{\mathrm{max}}}{e^{\lambda T_{\mathrm{min}}}-1} > 0.
    \end{split}
    \end{align}
    For any $\xi^x_{0,k} \in \mathcal{X}$, the evolved initial DCM by the foot placement $(L_k,T_k)$ is $\xi^x_{0,k+1} = \xi^x_{0,k} e^{\lambda T_k} - L_k$. Then, we can choose $(L_k,T_k)$ as boundary values $(L_{\mathrm{min}}, T_{\mathrm{min}})$ or $(L_{\mathrm{max}}, T_{\mathrm{min}})$, and bound $\xi^x_{0,k+1}$ as follows:
    \begin{align}
    \begin{split}
        \xi^x_{0,k+1} &\ge (\frac{L_{\mathrm{min}}}{e^{\lambda T_{\mathrm{min}}}-1}) e^{\lambda T_{\mathrm{min}}} - L_{\mathrm{min}} = \xi^x_{0,\mathrm{min}}, \\
        \xi^x_{0,k+1} &\le (\frac{L_{\mathrm{max}}}{e^{\lambda T_{\mathrm{min}}}-1}) e^{\lambda T_{\mathrm{min}}} - L_{\mathrm{max}} = \xi^x_{0,\mathrm{max}},
    \end{split}
    \end{align}
    i.e., $\xi^x_{0,k+1}$ remains in $\mathcal{X}$.
\end{proof}

A similar result also applies to the $y$-direction, where there exist two consecutive foot placements such that any initial DCM $\xi^{yl/r}_{0,k}$ has a bounded evolution. The bound of such $\mathcal{X}$ is referred to as the $\infty$-step capturability bound $d_{\infty}$ (or viability bound) in~\cite{koolen2012capturabilitybased} that theoretically distinguishes the stable walking gaits from the unstable ones, and such a bounded (i.e., viable) evolution of the DCM represents a walking gait that can be stabilized. Indeed, it can be shown that any evolution of the initial DCM in the interior of ${\mathcal{X}}$ can increase the viability margin, which implies an ``asymptotically stable'' walking system~\cite{wieber2002stability}.

For bipedal robotic walking on flat ground without terrain constraints, the allowable sets of each foot placement are sufficiently large and almost identical (though symmetric in the $y$-direction), thus we can always find at least one stable gait that converges to a desired periodic one given an appropriate initial DCM $\boldsymbol{\xi}_{0,1}$. However, even if the allowable sets of foot placements are constrained and the robot state is perturbed, as long as there exists an evolution of the initial DCM that ensures each foot placement lies within its corresponding allowable set, it is still possible to find a viable though aperiodic gait with the bounded DCM. This intuitively inspires a foot placement strategy on restricted footholds with force perturbations.

\section{Swing Foot Placement Planning via Discrete Model Predictive Control} \label{sec:hlc}

The primary contribution of this work is to design a foot placement planner for stepping stones with high accuracy of step positions and enhanced resistance to external perturbations. The stepping stones can be modeled as a sequence of bounds on each step position, which leads to significant constraints on the dynamic walking. Therefore, multiple future steps must be considered to find a stable walking gait by exploiting the viable evolution of the initial DCM with variable step duration.

To simplify the denotation, we define $\mathbf{z}_k:=\boldsymbol{\xi}_{0,k+1}$, $\mathbf{u}_k:=\mathbf{p}_{TD,k}$, and $\sigma_k:=e^{\lambda T_k}$, then the step-to-step discrete dynamics of the initial DCM by \eqref{eq:dcm_reset_map} is given by
\begin{align} \label{eq:dcm_discrete_dynamics}
     \mathbf{z}_k = \sigma_k \mathbf{z}_{k-1} - \mathbf{u}_k,\quad k=1,2,...\quad,
\end{align}
where $\mathbf{z}_0=\boldsymbol{\xi}_{0,1}=\boldsymbol{\xi}(0)$ is the initial DCM of the current step for $k=1$.

We now propose a discrete MPC formulation based on the discrete dynamics \eqref{eq:dcm_discrete_dynamics} that minimizes the errors between the gait parameters $(\mathbf{z}_k, \sigma_k, \mathbf{u}_k)$ and their desired values for multiple future steps. The design of the optimization constraints and objectives on the decision variables is as follows:

\subsubsection{\textbf{Step Position}}
For the Digit robot model with underactuated ankles and flat feet, we define the step position $\mathbf{u}_k$ as the geometric center at the bottom of each stance foot. A stepping stone profile is given as a sequence of fixed positions in the world frame, while the robot states and variables, such as the DCM and the step position, are defined in the associated contact frame. Given the preview of next $N$ stepping stones in the current contact frame, the target is to minimize the error between each future step position $\mathbf{u}_k$ and its desired step position $\mathbf{u}_{\mathrm{des},k}$ for $k=\{1,2,...,N\}$, which is computed as
\begin{align} \label{eq:u_k_des}
    \mathbf{u}_{\mathrm{des},k} = 
    \begin{cases}
        \mathbf{p}_{stone,1},\quad\quad\quad\quad\quad\quad\ \  k=1 \\
        \mathbf{p}_{stone,k} - \mathbf{p}_{TDC,k-1}, \quad k=2,...,N
    \end{cases}.
\end{align}

Since each $\mathbf{u}_k$ is defined in its corresponding contact frame, $\mathbf{p}_{TDC,k-1}$ should be the actual step position of the future ($k$-1)-th step in the current contact frame. Given that $\mathbf{p}_{TDC,k-1}$ are unknown during the current step, we use the previous planned values to approximate their actual values, which are iteratively updated inside the planner. In practice, if the planner has not returned the first result right after a touchdown, we then use the value of the ($k$-1)-th stone position $\mathbf{p}_{stone,k-1}$ instead.

Moreover, the bound on each step position is defined as a rectangular region around each $\mathbf{p}_{stone,k}$ that represents the physical dimension of each stepping stone.

\subsubsection{\textbf{Step Duration}}
Since we emphasize the accuracy of the step position and viability of the walking gait, we make the step duration a slack variable to relax the problem. We choose a constant nominal value $T_{k,\mathrm{des}}=T_{\mathrm{nom}}$ and a constant bound $[T_{\mathrm{min}}, T_{\mathrm{max}}]$ for $T_k$, and assign a small weight for the step duration term $\sigma_k$ in the MPC objective function.

\subsubsection{\textbf{Initial DCM}} \label{sec:initial_dcm}
The desired value of the initial DCM $\mathbf{z}_k$ is computed using \eqref{eq:dcm_xy_periodic} by substituting the desired values of the step position and duration of the $k$-th step. The bounds of the initial DCM are obtained using the mechanical limits of the robot model, which rules out all the infeasible states that lead to the robot falling. However, these bounds alone do not guarantee foot placements that can yield viable states for all the given stepping stones. Therefore, we choose an appropriately high weight in the MPC objective term for each $\mathbf{z}_k$ to track the desired value. This works as a soft constraint to enforce a viable evolution of the initial DCM through previewed stepping stones.

Moreover, since this foot placement planner runs iteratively during the swing phase, we use the measured value of the instantaneous DCM $\boldsymbol{\xi}_{\mathrm{mea}}(\tilde{t})$ to estimate the initial DCM of the current step $\mathbf{z}_0$ by computing it backward using \eqref{eq:dcm_solution}:
\begin{align} \label{eq:dcm_initial_value]}
    \mathbf{z}_0 = \boldsymbol{\xi}_{0,1} = \boldsymbol{\xi}_{\mathrm{mea}}(\tilde{t}) e^{-\lambda \tilde{t}},
\end{align}
where $\tilde{t} \in [0,T_1]$ is the current relative time of the step so that $\mathbf{z}_0$ is also iteratively updated using the robot states.

\subsubsection{\textbf{Discrete MPC Formulation}}
Combining \eqref{eq:dcm_discrete_dynamics}, \eqref{eq:u_k_des}, \eqref{eq:dcm_initial_value]}, and other constraints defined above, we now formulate the discrete MPC problem as follows:
\begin{align}
\begin{split}
    \underset{\substack{\mathbf{z}_k, \sigma_k, \mathbf{u}_k}}{\mathrm{arg\ min}}
        \quad \sum_{k=1}^{N} \beta_k \biggl(& \alpha_{z,k} \norm{\mathbf{z}_k - \mathbf{z}_{k,\mathrm{des}}}^2 \\
        + & \alpha_{\sigma,k} |\sigma_k-\sigma_{k,\mathrm{des}}|^2 \\
        + & \alpha_{u,k} \norm{\mathbf{u}_k-\mathbf{u}_{k,\mathrm{des}}}^2
        \biggr)\quad \\
\end{split} \\
    \mathrm{s.\ t.}\quad
    & \mathbf{z}_k = \sigma_k \mathbf{z}_{k-1} - \mathbf{u}_k, \\
    & \begin{bmatrix}
        z^x_{\mathrm{min}} \\ z^{yl/r}_{\mathrm{min}}
    \end{bmatrix} \leq \mathbf{z}_k \leq \begin{bmatrix}
        z^x_{\mathrm{max}} \\ z^{yl/r}_{\mathrm{max}}
    \end{bmatrix}, \\
    & \quad e^{\lambda T_{\mathrm{min}}} \leq \sigma_k \leq e^{\lambda T_{\mathrm{max}}}, \\
    & \begin{bmatrix}
        u^x_{k,\mathrm{min}} \\ u^{yl/r}_{k,\mathrm{min}}
    \end{bmatrix} \leq \mathbf{u}_k \leq \begin{bmatrix}
        u^x_{k,\mathrm{max}} \\ u^{yl/r}_{k,\mathrm{max}}
    \end{bmatrix}.
\end{align}
where $k=\{1,2,...,N\}$, and we use the weights $\alpha_{(\cdot),k}$ (with subscripts $z, \sigma, u$) and $\beta_k$ to represent soft constraints on the decision variable such that: $\alpha_{(\cdot),k}$ assigns relative weights on $(\mathbf{z}_k, \sigma_k, \mathbf{u}_k)$ for the $k$-th future step as we discussed above; $\beta_k$ is a decaying weight on each future step, addressing the importance of previewing earlier future steps, especially the imminent next one.

Although the discrete dynamics of DCM \eqref{eq:dcm_discrete_dynamics} takes a bilinear form about the decision variables $(\mathbf{z}_k, \sigma_k, \mathbf{u}_k)$, such a discrete formulation significantly reduces the problem dimension compared to a continuous-time MPC. Furthermore, we can utilize the latest commercial solvers to achieve online implementation.

\begin{remark}
    This MPC formulation is not limited to a particular terrain type. Appropriate weights and bounds to represent diverse control objectives, such as more accurate step position or more resistance to perturbation, can be designed and implemented in a variety of flat-ground robotic walking scenarios. Notably, both objectives can be interpreted as equivalent constraints on the evolution of the DCM. 
\end{remark}

\section{Adaptation of Variable Step Duration in the Low-level Controller} \label{sec:llc}


\begin{figure}[t]
    \centering
    \vspace{2mm}
    \includegraphics[width=\linewidth]{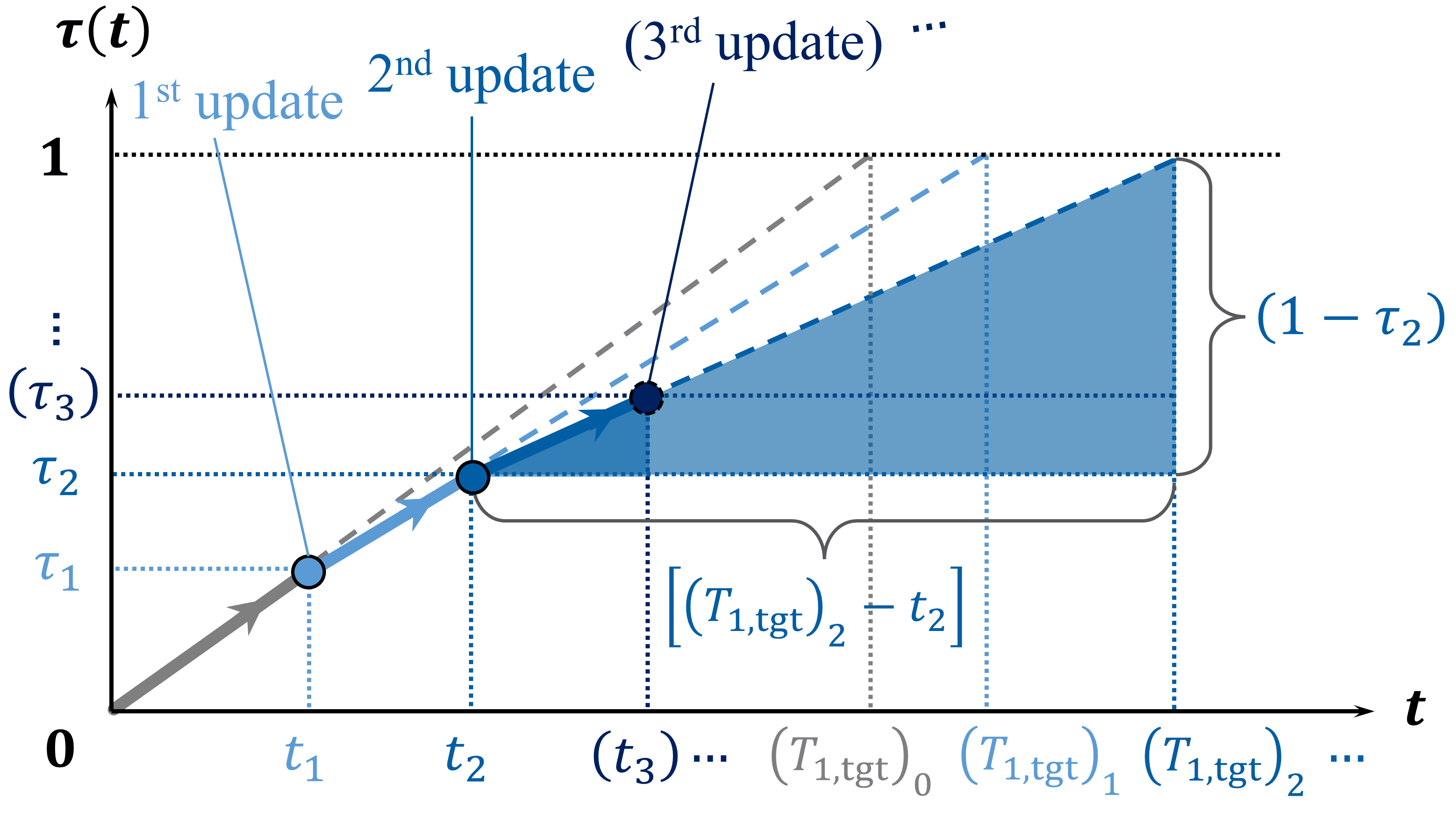}
    \caption{Illustration of updating $\tau(t)$ with respect to $(T_{1,\mathrm{tgt}})_i$. For instance, when $(T_{1,\mathrm{tgt}})_2$ is updated at $t_2$, $\tau(t)$ transitions linearly from $\tau_2$ to $1$ if $t$ reaches $(T_{1,\mathrm{tgt}})_2$, until $t$ reaches $t_3$ when the 3rd update takes place.}
    \label{fig:tau_update}
    \vspace{-2mm}
\end{figure}

The foot placement planner in \secref{sec:hlc} updates the target step position $\mathbf{p}_{TD,1,\mathrm{tgt}}$ and duration $T_{1,\mathrm{tgt}}$ of the current step at a low frequency during the current swing phase, while the low-level controller aims to regulate the robot's CoM dynamics towards the ALIP model and track a desired swing foot trajectory characterized by $(\mathbf{p}_{TD,1,\mathrm{tgt}}, T_{1,\mathrm{tgt}})$ at a high frequency. In this work, we use an existing whole-body controller presented in~\cite{castillo2023template} to track these task space outputs and maintain a stable walking gait.
The outputs of the low-level controller are defined using the robot configuration $q$ as
\begin{align}
    \mathcal{Y}(q) := \begin{pmatrix}
        \text{ base height } \\
        \text{ torso orientation } \\
        \text{ swing foot position } \\
        \text{ swing foot orientation } \\
    \end{pmatrix},
\end{align}
where the desired swing foot trajectory is generated as a 3-D polynomial curve ending at the desired step position with the desired step duration.

Note that the low-level controller uses a dimensionless phase variable $\tau$ to drive the trajectories of the robot, which increases monotonically from $0$ to $1$ during the current swing phase. If the desired step duration is constant, $\tau(t)$ is a linear function of the relative time $t$ with a fixed slope. However, when the desired step duration is variable during the swing phase, the phase variable $\tau$ needs to be computed accordingly to reflect the time-varying trajectory.
 
Let $n$ be the number of times the planner updates the desired foot placement during a swing phase, $t_i$ and $\tau_i$ ($i=\{0,1,...,n\}$) be the relative timestamp and the phase variable at the $i$-th planner update. As the swing phase starting from $t_0=0$ and $\tau_0 = 0$, the phase variable $\tau(t)$ is computed as:
\begin{align}
    \tau(t) = \tau_i + (1 - \tau_i) \frac{t - t_i}{(T_{1,\mathrm{tgt}})_i - t_i},\quad t\in [t_i, t_{i+1}], \label{eq:tau_t}
\end{align}
where $(T_{1,\mathrm{tgt}})_i$ is the $i$-th updated tgtired step duration. This design is primarily based on the following key points (See \figref{fig:tau_update}):
\begin{enumerate}
    \item $\tau(t)$ increases monotonically from $0$ to $1$ as $t$ increases from $0$ to the last updated desired step duration  $(T_{1,\mathrm{tgt}})_n$;
    \item $\tau(t)$ is continuous over $t$ and piecewise linear in all $[t_i, t_{i+1}]$;
    \item In each $[t_i, t_{i+1}]$, $\tau(t)$ increases with a fixed slope such that it starts from $\tau_i$ and aims to reach $1$ if $t$ reaches $(T_{1,\mathrm{tgt}})_i$, until the slope is updated with the new $(T_{1,\mathrm{tgt}})_{i+1}$ at $t=t_{i+1}$.
\end{enumerate}

Moreover, the time derivative of the phase variable $\dot{\tau}(t)$ is used to compute the time derivatives of the desired outputs $\dot{\mathcal{Y}}_{\mathrm{tgt}}(q)$ inside the low-level controller, which can be computed from \eqref{eq:tau_t} as:
\begin{align}
    \dot{\tau}(t) = \frac{1 - \tau_i}{(T_{1,\mathrm{tgt}})_i - t^i},\quad t\in [t_i, t_{i+1}].
\end{align}

\section{Simulation Results} \label{sec:results}
The proposed foot placement planner is tested on the humanoid robot model Digit in the MuJoCo simulation environment~\cite{todorov2012mujoco} on a personal laptop with an Intel i7 processor. The simulation runs at a fixed timestep of $1$ ms, the same as the low-level controller. The foot placement planner runs every $20$ ms, where the nonlinear MPC is solved by a state-of-the-art solver FORCES PRO~\cite{forcesnlp2017zanelli} (with average solving time around 5 ms). The number of future steps previewed in the MPC is chosen as $N=4$ to include two complete cycles of 3-D bipedal walking, though values up to 8 do not significantly increase the computation time. For all presented simulations, the parameters and constants of the foot placement planner are chosen as follows:
\begin{itemize}
    \item The MPC objective emphasizes enforcing the nominal initial DCM with accurate step position, thus the weights for the $k$-th future step are chosen as $(\alpha_{z,k}, \alpha_{\sigma,k}, \alpha_{u,k}) = (1\mathrm{e}4, 1, 2\mathrm{e}4)$, and $\beta_k=10^{4-k}$ for $k=\{1,2,3,4\}$, where we double the weights in the $y$-direction because the step position bound in $y$-direction is half of that in $x$-direction;
    \item The step position bound is defined as a $0.2\times0.1$ meters rectangular region centered at each $\mathbf{p}_{stone,k}$;
    \item The step duration has a nominal value $T_{\mathrm{nom}} = 0.5$ with the bound $[T_{\mathrm{min}}, T_{\mathrm{max}}]=[0.3, 0.7]$ seconds;
    \item The initial DCM bounds are computed using the mechanical limits of the robot such as $|p^x_{TD,k}| \le 0.6$ and $0.1 \le |p^y_{TD,k}| \le 0.5$ meters with the step duration bound above.
\end{itemize}

For a comprehensive test of the planner performance, four types of stepping stone profiles and a perturbation scenario with four different perturbations are designed to represent the challenging terrain on flat ground, where we compare our $N$-step preview planner (denoted as \textbf{DCM-MPC}) with the one-step preview planner in~\cite{khadiv2020walking} (denoted as \textbf{DCM-QP}). 

\begin{table}[t]
    \vspace{2mm}
    \caption{Stepping Stones Profiles}
    \label{tab:stepping_stone_profile}
    \centering
    \renewcommand{\arraystretch}{1.5}
    \begin{tabular}{ccc}
        \hline
        Profile No. & $L_j$ (m) & $W_j$ (m) \\
        \hline
        I & fixed at $\{0.2,0.4,0.6\}$ & fixed at $\{-0.15,0.15\}$ \\
        II & $U[0.2, 0.5]$ & $U[-0.15, 0.15]$ \\
        III & $(0.2$*$8,0.5$*$8)$*$2$ & $U[-0.15, 0.15]$ \\ 
        IV & $U[0.2, 0.5]$ & $(-0.1$*$8,0.1$*$8)$*$2$ \\
        \hline
    \end{tabular}
    \begin{tablenotes}
        \item \emph{Profile I} contains $3\times2$ combinations of fixed $L_j$ and $W_j$; \emph{Profile II$\sim$IV} includes one or two lists of uniformly distributed random values as $L_j$ and/or $W_j$, denoted as $U[\cdot, \cdot]$; \emph{Profile III} uses a list of alternating values as $L_j$, which is 8 consecutive values of $0.2$ followed by 8 values of $0.5$ and repeated once, which emphasizes on abrupt changes of stone positions in the $x$-direction (similar to $W_j$ in \emph{Profile IV}).
    \end{tablenotes}
\end{table}

\subsection{Robust Walking on Stepping Stones}

The stepping stone profiles are listed in \tabref{tab:stepping_stone_profile}. Each profile contains 32 stones with each relative position computed using $(L_j,W_j)$ as $\mathbf{p}_{stone,\mathrm{rel},j}=[L_j, w^{l/r}+W_j]^T$ ($j=\{1,...,32\}$), where $w^{l/r}$ is chosen as $\pm 0.28$ meters.

In the simulation, the robot is initialized in a standing posture with zero velocity and then walks 6 steps to approach the first stepping stone. In this speeding-up stage, the robot is only commanded to reach an appropriate $x$-velocity before walking onto the stepping stones and thus may run into different initial conditions for the stepping stone test. Though such uncertainty helps highlight the robustness of our planner, we try to ensure consistent initial conditions in each comparison test between DCM-MPC and DCM-QP. As we will discuss below, DCM-MPC outperforms DCM-QP in all the presented simulation results.

\begin{table}[t]
    \caption{Step Position Errors in Profile I}
    \label{tab:profile_i}
    \centering
    \renewcommand{\arraystretch}{1.5}
    \begin{tabular}{c|cc|cc|cc}
        \hline
        \diagbox[width=50pt, height=16pt, innerleftsep=2pt, innerrightsep=2pt]{$W_j$}{$L_j$} & \multicolumn{2}{c|}{$0.2$} & \multicolumn{2}{c|}{$0.4$} & \multicolumn{2}{c}{$0.6$} \\
        \hline
        $0.15$ & $\mathbf{0.011}$ & $0.018$ & $\mathbf{0.021}$ & 0.027 & $\mathbf{0.025}$ & N/A \\
        $-0.15$ & $\mathbf{0.014}$ & $0.034$ & $\mathbf{0.028}$ & N/A & $\mathbf{0.030}$ & N/A \\
        \hline
    \end{tabular}
    \begin{tablenotes}
        \item Comparison of step position errors (in meters) in the test of Profile I (left and bold: DCM-MPC, right: DCM-QP). DCM-MPC outperforms DCM-QP in these tests, where ``N/A'' indicates that DCM-QP fails.
    \end{tablenotes}
\end{table}

\subsubsection{Profile I} \tabref{tab:profile_i} lists the step position errors that are computed as the root-mean-square errors (RMSE) between the actual step positions and stone positions. DCM-MPC completes all the tests successfully with small step position errors, while DCM-QP is more likely to fail when $L_j$ takes large values such as $0.6$ meters (which is close to the mechanical limit of the robot). Note that the aforementioned initial conditions are not symmetric for the tests with $W_j= 0.15$ and $-0.15$ meters, where the latter seems to be more challenging because the first stepping stone in the tests usually requires the right swing foot. DCM-QP fails to maintain a viable gait starting from the first step onto the stepping stone due to such harsh initial conditions, while DCM-MPC benefits from the strategy that considers the viability of multiple future steps to decide the next foot placement.

\begin{figure}[t]
    \vspace{2mm}
    \centering
    \includegraphics[width=\linewidth]{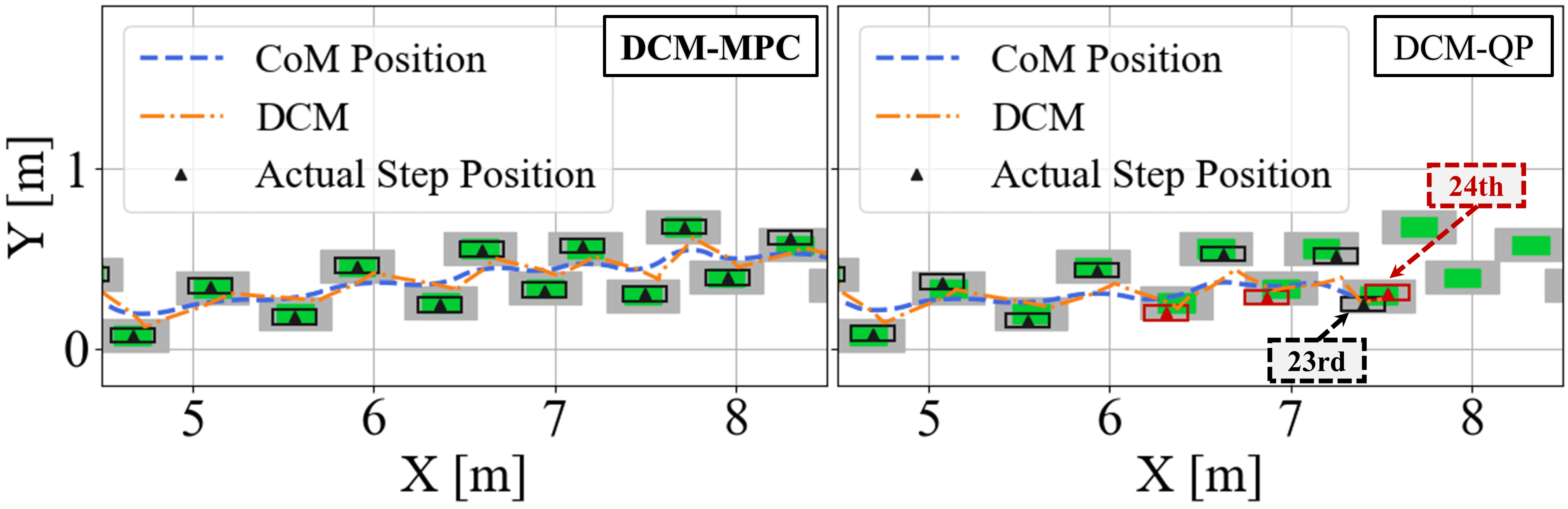}
    \caption{Comparison of footprints in Profile II test. The black triangles mark the actual step positions, with the black frames illustrating the flat feet of the Digit, while the green rectangles mark the step position bounds on the gray stepping stones. (Left, DCM-MPC) The robot walks stably through the random stones with the bounded DCM; (Right, DCM-QP) The robot falls after the 24th step (marked as red frames) due to a non-viable foot placement of the 23rd step.}
    \label{fig:profile_ii_1}
\end{figure}

\begin{figure}[t]
    \vspace{2mm}
    \centering
    \includegraphics[width=\linewidth]{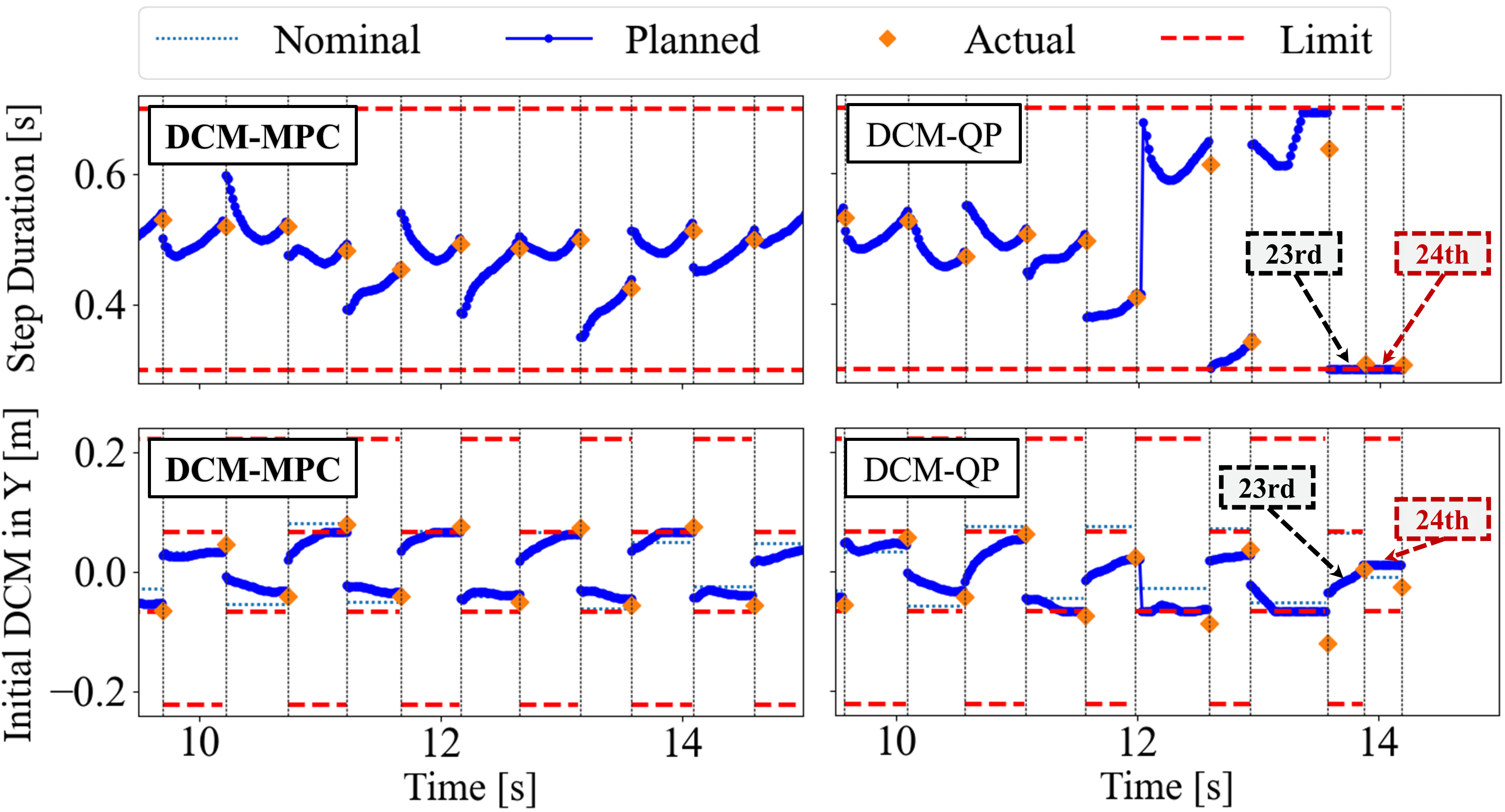}
    \caption{Comparison of step duration and initial DCM in the test of Profile II. The vertical gray lines mark the touchdown moments of walking steps. (Left, DCM-MPC) The step duration is adjusted to bound the DCM with smoother changes; (Right, DCM-QP) DCM-QP fails with a boundary step duration that indicates the non-viable DCM.}
    \label{fig:profile_ii_2}
\end{figure}

\begin{figure}[t]
    \vspace{2mm}
    \centering
    \includegraphics[width=\linewidth]{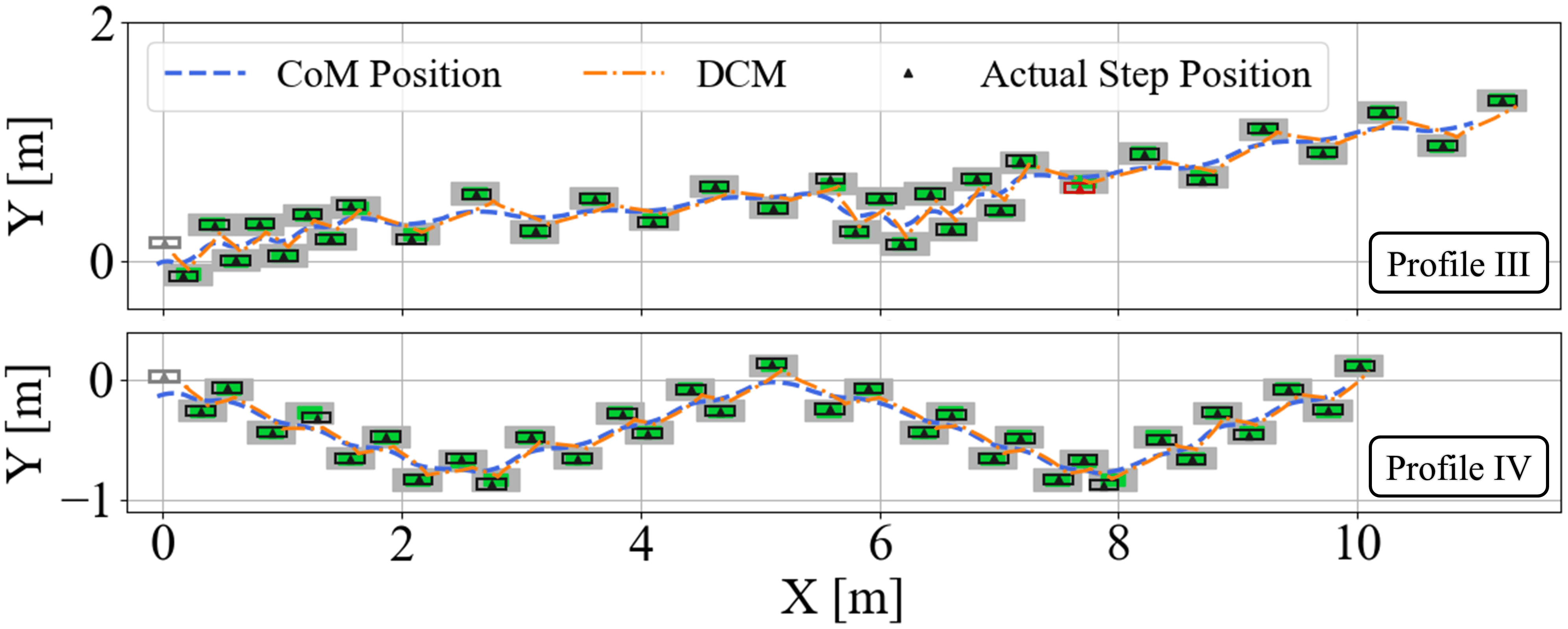}
    \caption{Footprints of DCM-MPC in the tests of Profile III (Upper) and Profile IV (Lower). The robot completes these tests without any step violating the step position constraint.}
    \label{fig:profile_iii_iv}
\end{figure}

\begin{figure}[t]
    \centering
    \vspace{2mm}
    \includegraphics[width=\linewidth]{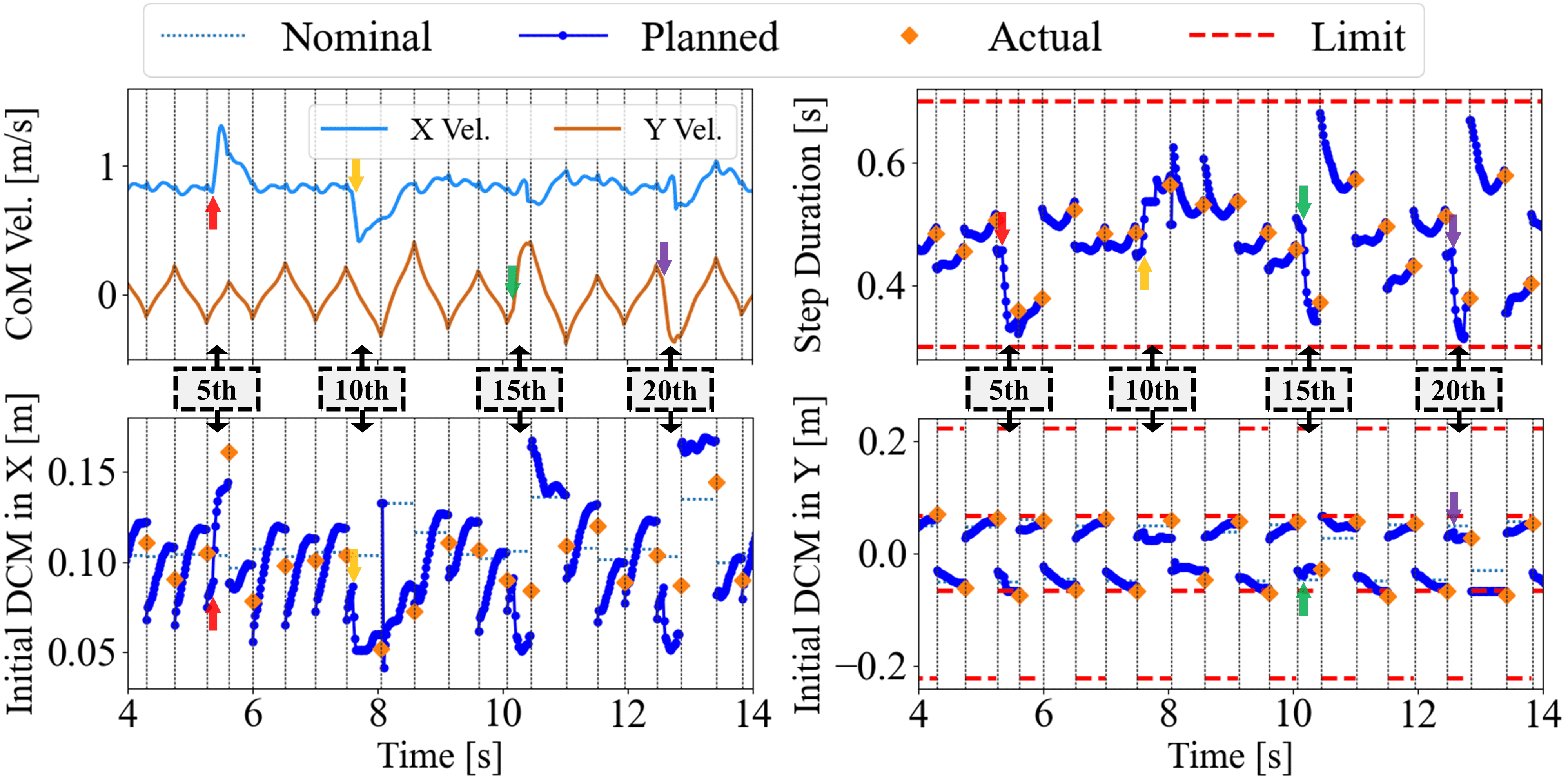}
    \caption{Robot states of DCM-MPC in the perturbation test. Four perturbation forces are marked by red, yellow, green, and purple arrows, respectively. (Upper Left) The $x$- and $y$-velocity change abruptly due to the perturbations but are quickly restored; (Upper Right) The step duration is adjusted in response to the perturbations to maintain viable states; (Lower Left and Right) The initial DCM in both $x$- and $y$-directions are bounded.}
    \label{fig:profile_p}
\end{figure}

\subsubsection{Profile II} Each relative stone position is given by randomly generated $(L_j, W_j)$. DCM-MPC completes this test with a step position error of $0.029$ meters, while DCM-QP ends halfway with the robot falling. \figref{fig:profile_ii_1} compares the footprints, the CoM, and the DCM trajectories of two planners. DCM-QP fails to place its foot onto the desired 24th stone because the 23rd foot placement yields an initial DCM that cannot lead to viable states given the 24th stone position, highlighting the significance of the multi-step preview. \figref{fig:profile_ii_2} compares the step duration and initial DCM in $y$-direction of two planners, where DCM-MPC can find foot placements to maintain viable states with smoother changes in the actual step duration.

\subsubsection{Profile III and IV} These two profiles emphasize the abrupt changes in the $L_j$ and $W_j$, respectively. The step position errors of DCM-MPC in these two tests are $0.025$ and $0.026$ meters, while DCM-QP falls after a few steps. \figref{fig:profile_iii_iv} shows the footprints of DCM-MPC in both tests. Even under such challenging stepping stone profiles, especially with large variations in stone positions in the lateral direction, DCM-MPC can find foot placements to maintain viable states.

\subsection{Perturbation Test on Stepping Stones}

The test scenario with perturbations contains 24 steps on the stepping stones profile $(L_j,W_j)=(0.4,0)$ meters with the step width bound relaxed by $0.1$ meters and four force perturbations applied from all four directions to the pelvis of the Digit robot. Each perturbation force lasts from the relative time $t=0.1$ to $0.2$ seconds in the specified step with the forces and directions as follows: 1) 5th step, $F^x=175$\,N; 2) 10th step, $F^x=-175$\,N; 3) 15th step, $F^y=125$\,N; 4) 20th step, $F^y=-125$\,N.
The robot states during the perturbation test are plotted in \figref{fig:profile_p}. Since we choose a stone profile where the robot walks straightforwardly, the perturbation resistance of both planners in the $x$-direction is similar. During the first two perturbations, our approach demonstrates excellent capability of handling adversarial disturbances, where the step duration of the current step is rapidly adjusted to keep balance. 
When perturbation directions are orthogonal to the walking direction (i.e., in the $y$-direction), the feasibility of viable steps is significantly reduced. The baseline DCM-QP approach fails under such disturbances, whereas the DCM-MPC remains stable by considering a longer horizon of locomotion, and hence capable of regulating the durations of multiple future steps to improve stability and robustness.

\section{Conclusion} \label{sec:conclusion}

This paper presents a discrete MPC formulation for multi-step preview foot placement planning based on the DCM and capturability analysis. We demonstrate its performance in achieving more robust bipedal robotic walking on challenging terrains with highly restricted footholds and severe perturbations, which benefits from variable step duration.
Compared to other MPC formulations for stepping stones, the discrete dynamics of the initial DCM vastly simplify the optimization problem, thus ensuring real-time implementation.
Future work will focus on validating its effectiveness in real-world experiments and further analyzing gait robustness on uneven terrains. 
\vspace{1em}
\balance
\bibliographystyle{IEEETran}
\bibliography{references}

\end{document}